\documentclass[10pt,compsocconf,conference]{IEEEtran}
\IEEEoverridecommandlockouts
\newcommand{\ApproxSign}{\raise.17ex\hbox{$\scriptstyle\sim$}}

\usepackage{cite}
\usepackage{lipsum}
\usepackage{multirow}
\usepackage{amsmath,amssymb,amsfonts}
\usepackage{algorithmic}
\usepackage{graphicx}
\usepackage{textcomp}
\usepackage{float}
\usepackage{blindtext}
\usepackage{xcolor}
\usepackage{caption}
\usepackage{array}
\usepackage{booktabs}
\newcolumntype{C}[1]{>{\centering\let\newline\\\arraybackslash\hspace{0pt}}m{#1}}

\def\BibTeX{{\rm B\kern-.05em{\sc i\kern-.025em b}\kern-.08em
    T\kern-.1667em\lower.7ex\hbox{E}\kern-.125emX}}

\newcommand{\NJ}[1]{{}}

\newcommand\blfootnote[1]{%
  \begingroup
  \renewcommand\thefootnote{}\footnote{#1}%
  \addtocounter{footnote}{-1}%
  \endgroup
}

\begin{document}

\title{The Ramifications of Making Deep Neural Networks Compact\\
{}
}
\author{\IEEEauthorblockN{Nandan Kumar Jha, Sparsh Mittal, Govardhan Mattela}
\IEEEauthorblockA{
Department of Computer Science and Engineering, IIT Hyderabad, India\\ 
%\\Oak Ridge, Tennessee, USA\\
Email: \{cs17mtech11010, sparsh, cs17resch01004\}@iith.ac.in}
}

\maketitle

\begin{abstract}

The recent trend in deep neural networks (DNNs) research is to make the networks more compact. The motivation behind designing compact DNNs is to improve energy efficiency since by virtue of having lower memory footprint, compact DNNs have lower number of off-chip accesses which improves energy efficiency. 
However, we show that making DNNs compact has indirect and subtle implications which are not well-understood. Reducing the number of parameters in DNNs increases the number of activations which, in turn, increases the  memory footprint. We evaluate several recently-proposed compact DNNs on  Tesla P100 GPU and show that their ``activations to parameters ratio'' ranges between 1.4 to 32.8. Further, the `` memory-footprint to model size ratio'' ranges between 15 to 443.
This shows that a higher number of activations causes large memory footprint which increases on-chip/off-chip data movements. Furthermore, these parameter-reducing techniques reduce the arithmetic intensity which increases on-chip/off-chip memory  bandwidth requirement. Due to these factors, {\em the energy efficiency of compact DNNs may be significantly reduced which is against the original motivation for designing compact DNNs}.   
\end{abstract}

\begin{IEEEkeywords}
Deep neural networks (DNNs), compact DNNs, activations, parameters,  energy-efficiency, embedded systems. 
\end{IEEEkeywords}

\section{Introduction} \label{sec:introduction}
Deep neural networks (DNNs) have shown phenomenal results in various domains such as image classification and object detection, etc. \cite{AlexNet,VGG,GoogleNet}.
After the success of AlexNet \cite{AlexNet}, to improve accuracy, researchers have proposed even deeper \cite{VGG,he2016identity} and wider  \cite{ResNext,InceptionV2,InceptionV4} networks which are deemed as over-parameterized\blfootnote{Support for this work was provided by Science and Engineering Research Board (SERB), India, award number ECR/2017/000622.}. These networks have huge compute, memory and power demands which hinders their deployment on resource-constrained embedded and mobile devices \cite{ref102}.
To enable the deployment of DNNs on resource-constrained platforms, researchers have proposed two types of   heuristics: (1) compressing the existing over-parameterized deeper and wider networks and (2) designing new algorithms which have very few  parameters i.e. compact models. For example, Han et al. \cite{DeepCompression} propose magnitude-based pruning of filter weights and Yang et al. \cite{EPruning} propose energy-aware pruning to improve energy efficiency. These pruning methods, however, require exhaustive retraining to achieve the accuracy of the original pre-trained model.

\begin{figure}[htbp]
\begin{center}
\fbox{\includegraphics[scale=0.5]{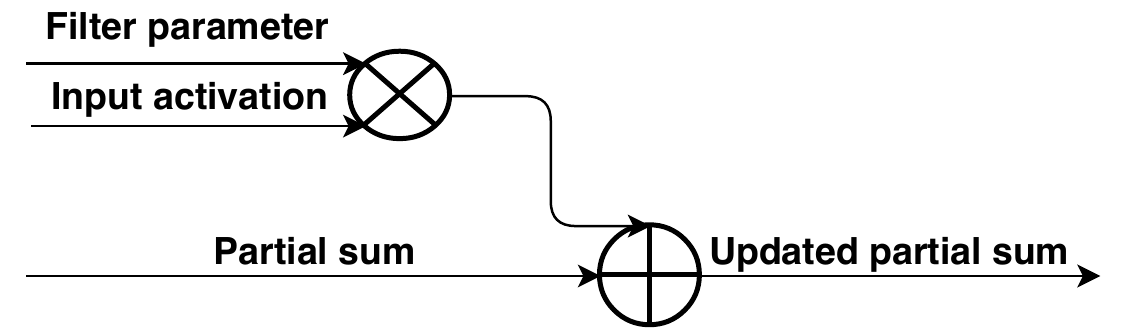}}
\caption{Illustration of a MAC operation. }
%\caption{One MAC  involves 3 reads as filter parameter, input activation partial sum and one write as update partial sum.}
\label{fig:MAC}
\end{center}
\end{figure}

Compact DNNs have the advantage that they avoid retraining overheads and after training, they  can be directly deployed on resource constrained devices. The current trends in designing compact DNNs is to reduce the number of parameters and computations by leveraging error-tolerance of DNN application domains. Reducing the number of parameters helps in fitting the network in limited on-chip memory and avoids expensive off-chip access. This makes the DNNs energy efficient. The computational cost of a DNN is measured in terms of number of MAC (multiply-accumulate) operations performed in conv and FC layers. %Number of parameters and MACs are the proxy metrics for memory footprint and total energy consumption, respectively  \cite{Netadapt}.
Since measuring the memory footprint and total energy consumption is not straightforward, researchers generally use number of parameters and number of MACs (respectively) as their proxies \cite{Netadapt}. However, this approach has crucial limitations. The total memory footprint is sum of the (1) size of weights (2) size of activations and (3) gradients corresponding to activations and parameters. Hence it  depends on both the number of parameters and activations as well. However, since the number of activations cannot be estimated from the number of parameters, the number of parameters is not a good indicator of memory footprint. Further,  as shown in Figure \ref{fig:MAC}, one MAC operation requires three read and one write operations. Hence, the energy consumed in each MAC operation depends on (1) the location of the operands in the memory hierarchy, such as register file, cache or main memory which decides the operand  fetch energy \cite{ref94} and (2) the type of convolution such as $3\times3$, $1\times1$ or depth-wise separable convolution which decides the degree of reuse \cite{Notallops}. Hence, the number of MAC is not an accurate indicator of the energy consumption of  a DNN. 

In this paper, we study state-of-the-art compact DNNs and analyze the unforeseen implications of reducing the number of parameters. For example,  1.0-G-SqNxt-23, a variant of SqueezeNext\cite{SqueezeNext} has $112\times$ fewer parameters than AlexNet (Table \ref{tab:Modelattributes}) but higher memory footprint than AlexNet (Table \ref{tab:ResultsSummary}). With larger batch size ($B$), this becomes even worse and for $B=100$,  1.0-G-SqNxt-23 consumes $5.7\times$ higher  memory compared to AlexNet. Further, 1.0-G-SqNxt-23 has $3.27\times $ fewer MACs than AlexNet (Table \ref{tab:Modelattributes}) but the energy efficiency  of 1.0-G-SqNxt-23 is  $45\times$ lower  than that of AlexNet (Table \ref{tab:ResultsSummary}). On digging deeper to understand the sources of inefficiency, we found that 1.0-G-SqNxt-23 has $8.7\times$ higher activation and $28.5\times$ lower MACs/activation ratio compared to AlexNet. Lower MACs/activation ratio leads to lower arithmetic intensity and makes the DNN bandwidth bound. This increases the on-chip/off-chip memory access and results into higher energy consumption. 
%SqueezeNext \cite{SqueezeNext} use fire module in  more aggressive manner to reduce the number of parameters and has 112X fewer parameters than AlexNet while maintaining same accuracy as that of the AlexNet. MobileNets \cite{MobileNet} primarily optimized for latency but also make network compact. It  factorize original standard convolution into a depthwise separable convolution,which uses single filter for each input channel, and a 1X1 pointwise convolution to combine the outputs of the  depthwise separable convolutions. DenseNet \cite{DenseNet} optimize for feature propagation and feature reuse using desne connection between the layers of dense blocks. It also lessen the number of parameters by fixing the number of output feature maps (growth rate) and making use of 1X1 filters as bottleneck layers before transition layers. Similarly GoogleNet and InceptionV2 use 1X1 filters as bottleneck layers in inception module. This bottleneck layers reduce the number of input feature map before convolution with larger filters (3X3 or 5X5).
We summarize our contributions as follows. 
\begin{itemize}
\item We analyze DNNs which are representative of the state-of-the-art compact DNNs. We perform kernel-level analysis of compact DNNs to get insights into utilization of compute resources by the MACs and the performance bottlenecks of each DNN.

%, primarily streaming multiprocessor, which helps in understanding of  low throughputs of some of compact DNNs.

%(cuda/cuBLAS) level analysis of each compact DNNs using Nvidia profiler and also analyze critical kernels using Nvidia visual profiler. This analysis explain how MACs are utilizing the compute resources , primarily streaming multiprocessor, which helps in understanding of  low throughputs of some of compact DNNs.

\item We find  implications of making  DNNs compact  on memory footprint, energy efficiency and throughput.  We find that memory footprint depends not only on the number of parameters but activations also. In fact, the contribution of activations in memory footprint is very high.
%\item We find that memory footprint depends on both the number of parameters and activations also the number of activations has very high contribution in memory footprint.
\item Since measuring the  arithmetic  intensity is relatively difficult, we propose using  MACs/parameter ratio and MACs/activation ratio as the proxies for  them. Low arithmetic  intensity indicates lower degree of reuse of parameters and activations which increases energy required for processing inputs.  
\end{itemize}

\section{Ramifications of Making DNNs Compact} \label{sec:CDNN}
\textbf{Notations:}  Table \ref{tab:DNNshapeParameters} shows the terms used for measuring the dimension of DNN layers. Equations \ref{eqn:param}, \ref{eqn:act} and \ref{eqn:MACs} show the formula for computing the number of parameters, activations and MACs (respectively), assuming batch size of one \cite{Mobilenet}. 
%We use a fixed input size of  $224\times 224\times 3$ in all our experiments.

%Notice that the number of parameters depends on the filter size, number of filter channels (which is equal to number of input channel) and number of 3-D filters. Similarly, number of MACs  depends on the number of channels in input feature maps and output feature map,  filter size (2-D) and input size.  

\begin{table} [htbp] 
\caption{Dimension of layers in DNN}
\label{tab:DNNshapeParameters}
\centering
\begin{tabular}{ |c | c| } 
 \hline
 \textbf{DNN shape parameters } & \textbf{Description}  \\
 \hline 
 $N$ & $\#$ input (input feature map) channels \\
 $M$ & $\#$ output (output feature map) channels \\
 %$F$ & $\#$ filter (3-D)    \\
 $S_{N}$ & input (input feature map) width/height  \\
 $S_{M}$ & output (output feature map) width/height  \\
 $S_{F}$ & filter (2-D) width/height  \\
\hline
\end{tabular}
\end{table} 

\begin{align} 
\label{eqn:param}
\# Parameters &= N\times M\times S_{F}\times S_{F} \\
\label{eqn:act}
\# Activations &= M\times S_{M}\times S_{M} \\
\label{eqn:MACs}
\# MACs &= N\times M\times S_{M}\times S_{M}\times S_{F}\times S_{F}
\end{align}

\begin{figure}[htbp]
\centering
\fbox{\includegraphics[scale=0.6]{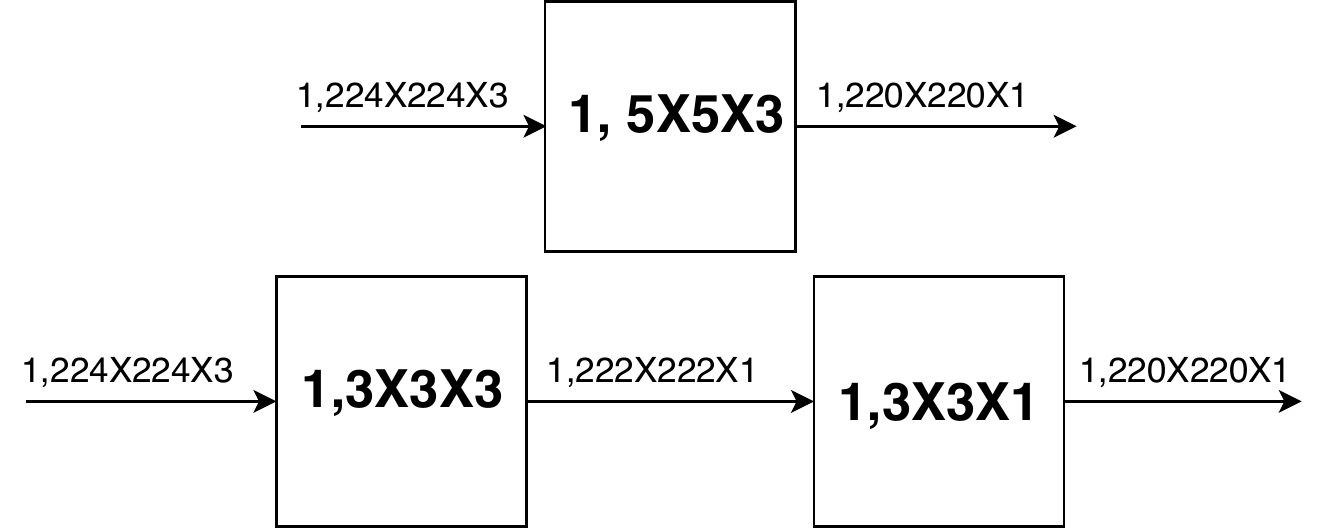}}
\caption{Factorizing one $5\times 5$ filter into two $3\times 3$ filters produces equal sized output feature map. }
\label{fig:ParameterActivationCalculation}
\end{figure}

\textbf{Filter factorization and its impact:} To reduce the number of parameters, larger size filter is factorized into equivalent smaller size filters. As shown in Figure \ref{fig:ParameterActivationCalculation}, factorizing one $5\times 5$ filter into two $3\times 3$ filter produces equal-sized output feature map (here in all convolution padding is zero and stride is one). We study the impact of this filter-factorization on the number of parameters, activations and MACs. We found that this factorization reduces the number of  parameters and MACs by 52\% and 51.3\% respectively, but increases the number of output activations by 102\%. This may lead to an increase in memory footprint in DNNs with high degree of nonlinearity (refer section \ref{sec:discussion}) even though the number of parameters is reduced  by half.

%Further, the additional layers between two $3\times3$  filters, such as BatchNorm and/or ReLU, increase the number of activation while the number of parameters remains the same. 
%We observe that on adding one ReLU layer between the two $3\times3$  filters shown in Figure \ref{fig:ParameterActivationCalculation} increases the total number of output activations by 204\%!
However, previous works  have ignored the impact of this increase on memory footprint and energy efficiency.  Although this factorization increases the  MACs/parameter ratio only by 1.34\% but decreases the MACs/activation ratio by 61\%. This leads to lower arithmetic intensity which, in turn, increases the memory accesses. {\em This reduces the overall energy efficiency}. Hence, contrary to the popular belief,  this parameter-reducing techniques degrade energy efficiency. 

%Notice that activation to parameter ratio of AlexNet is very low while that of the compact DNNs ranges from 1.44 to 34.07. Also, the MACs per parameter of compact DNNs are higher than AlexNet but MACs per activation is quite low compared to AlexNet.

\begin{table*} [htbp]
\caption{Characteristics of compact DNNs (Params and Acts refer to  parameters and activations of DNNs, respectively). }
\label{tab:Modelattributes}
\centering
\begin{tabular}{ |c| c| c| c| c| c| c| c | } 
 \hline
 \textbf{Model Name} & \textbf{Image size} & \textbf{MACs (M)} & \textbf{ $\#$Params (M)}  &  \textbf{$\#$Acts (M)}  & \textbf{Acts/Params} & \textbf{MACs/Params} & \textbf{MACs/Acts} \\
 \hline
 AlexNet             & $224\times 224$ &  723    & 60.97  &  2.05 & 0.03  &11.86  & 352.65    \\
 \hline
 SqueezeNet-V1.0     & $224\times 224$ & 848   & 1.25   & 12.3 & 9.84 &678.08 & 68.91    \\
 SqueezeNet-V1.1     & $224\times 224$ & 349    & 1.24   &  7.2 & 5.81  &281.57 & 48.49        \\
 \hline
 1.0-G-SqNxt-23      & $224\times 224$ & 221    & 0.54   & 17.81  & 32.80 &406.35 & 12.39     \\
 1.0-SqNxt-23        & $224\times 224$ & 273    & 0.72   & 17.81  & 24.84 &380.50 & 15.32     \\
 1.0-SqNxt-23v5      & $224\times 224$ & 225    & 0.93   & 14.06 & 15.12 &242.04 & 16.01        \\
 2.0-SqNxt-23        & $224\times 224$ & 726    & 2.36   & 32.21 & 13.65 &307.62 & 22.54       \\
 2.0-SqNxt-23v5      & $224\times 224$ & 703    & 3.22   & 24.66 & 7.66  &218.41 & 28.52       \\
 \hline
 1.0-MobileNet-224   & $224\times 224$ &  574   & 4.23   & 20.32 & 4.80  &135.65 & 28.24       \\
 DenseNet-121        & $224\times 224$ &  3080  & 7.98   & 69.99 & 8.77  &385.96 & 44.01        \\
 \hline
 GoogLeNet           & $224\times 224$ & 1590   & 7.00      & 10.06 & 1.44  &227.14 & 158.05      \\
 Inception-V2         & $231\times 231$ & 2200   & 11.19  & 18.03 & 1.61  &196.60 & 122.02       \\
 \hline
\end{tabular}
\end{table*}

We now discuss the architecture and building blocks of several compact DNNs. Table \ref{tab:Modelattributes} summarizes their characteristics. We also discuss the implications of the parameter-reducing techniques and highlight the sources of inefficiency.
 
\subsection{SqueezeNet} 
To reduce the number of parameters,  SqueezeNet \cite{SqueezeNet} uses fire module as a building block. Both SqueezeNet versions, V1.0 and  V1.1, have 8 fire modules (fire2 - fire9) which consist of squeeze and expand layers. 
The squeeze layer has only $1\times 1$ filters which serves as the bottleneck layer similar to that in the inception module\cite{GoogleNet} and reduces the dimensionality of input feature map before $3\times 3$ convolutions performed  in the expand module. It reduces the number of computation (MACs) and the parameters. The fire module has majority of  $1\times 1$ filters. Replacing a $3\times 3$ filter with a $1\times 1$ filter reduces the number of parameters by $9\times $. SqueezeNetV1.1 has $2.2\times $ fewer MACs than SqueezeNetV1.0.
%since it has 64, $3\times 3$ filters compared to 96, $7\times 7$ filters in SqueezeNetV1.0 in its first convolution layer. 

\textbf{Limitations:} In  squeeze layer of fire module, $1\times 1$ filter reduces the number of activations whereas in expand layer, it increases the number of activations. 
%The ReLU layers after the $1\times1$ convolution in both squeeze and expand layers and also after the $3\times 3$ convolution in expand layer do not alter the number of parameters but increase the number of activations significantly.
 Compared to AlexNet, SqueezeNet has $50\times$ fewer parameters and at $B=1$, it has lower memory footprint, however with increasing value of batch size, its memory footprint increases at a faster rate. For example, at $B=128$, AlexNet and SqueezeNetV1.0 consume 2287MB and 8225MB of global memory on GPU, respectively. 
%Although SqueezeNet has 50X fewer parameters than AlexNet and at $B=1$, it consumes lower memory t han but it consumes higher memory. For example, in our experiments, on using a batch size of 128, the g in GPU when batch size is 128 (8387MB vs 2305MB, as shown in Table \ref{tab:ResultsSummary}.
Also, using small filters lowers data reuse as well as arithmetic intensity  which makes the convolution operation bandwidth bound. By comparison, in large filters, convolution operation is computation-bound. Reducing the degree of data reuse lowers energy efficiency since the MAC operands are fetched from lower levels of  memories such as DRAM and not from upper levels such as L1 cache. 
 
\subsection{SqueezeNext} \label{sec:SqNxt}
%SqueezeNext \cite{SqueezeNext} uses the concept of fire module in more aggressive way and reduces the model size to 2.3MB. It uses two stage bottleneck module to reduce the number of input feature maps to the 3X3 convolution. Each bottleneck stage reduces the number of feature map by two. Instead of replacing 3X3 filters with 1X1 filters, it decomposes the 3X3 filters into 1X3 and 3X1 filters and gains  extra reduction of 1.5X in the number of parameters. It also uses skip  connection across the fire block to facilitate easy training since the  network becomes deeper with the use of two bottleneck stage and decomposed convolution. It further reduces the number of parameters by using a bottleneck layer before the FC (fully connected) layer. 1.0-G-SqNxt-23 (refer Table\ref{tab:Modelattributes}) is the baseline model and other networks are variants of it. The baseline model has total of 21 fire module blocks where each block has two bottleneck layers, two decomposed convolution layer and one expansion layer. 

In SqueezeNext family \cite{SqueezeNext}, 1.0-SqNxt-23 is the baseline model which has 23 building blocks, where each building block is an aggressive version of the fire block. Each fire block has (1) two stage bottleneck layers (2) decomposed $3\times 3$  (into $1\times 3$ and $3\times 1$) filters and (3) skip connection across the block. Each bottleneck stage reduces the number of output feature map by a factor of two and this reduces the number of input feature maps to $3\times3$ convolution in the expand layer. Decomposing a $3\times3$ filter into a $1\times3$ and a $3\times1$ filter leads to  an additional  reduction of $1.5\times$  in the number of parameters.  It further reduces the number of parameters by using a bottleneck layer before the fully connected layer \cite{SqueezeNext}. 

Among these 23 blocks, four groups of size 6 blocks, 6 blocks, 8 blocks and 1 blocks (from input to output) have input feature map resolution of $55\times 55$ (64 channels), $55\times 55$  (32 channels), $28\times 28$  (64 channels) and $14\times 14$  (128 channels), respectively. Blocks with higher input resolution lead to lower arithmetic intensity because they have higher number of activations. The distribution of blocks in 1.0-G-SqNxt-23, 1.0-SqNxt-23 and 2.0-SqNxt-23 \cite{SqueezeNext} are [6,6,8,1] whereas in the v5 versions (1.0-SqNxt-23v5 and 2.0-SqNxt-23v5) this is [2,4,14,1] \cite{SqueezeNext}. 
%Use of lower number of blocks in group with higher input resolution and higher number of block in group with lower input resolution increases the parameters but reduces activations and MACs.

\textbf{Limitations:} SqueezeNext uses a  two-stage bottleneck layer in each fire module and the number of such fire blocks is nearly $3\times$ higher  compared to that in SqueezeNet. This aggravates  the problem  of $1\times1$ convolution since it leads to low data reuse and low arithmetic intensity. The skip connection across a fire block further aggravates the problem of higher number of activations because it requires more feature maps in global memory of GPU. Although the 1.0-G-SqNxt-23 has $112\times $ fewer parameters than AlexNet but its memory footprint is higher than that of the AlexNet. As shown in Table \ref{tab:ResultsSummary}, at batch size one, the memory consumption of 1.0-G-SqNxt-23 and  AlexNet are 1019MB and  1015MB, respectively. With increasing batch size, the memory footprint of SqueezeNext increases rapidly. 

\subsection{MobileNet}
MobileNet \cite{Mobilenet} was designed to reduce the inference time. MobileNet reduces the computational complexity, i.e. number of MACs, by allowing only a single filter to convolve with each input feature map. MobileNet uses depthwise separable convolution which factorizes a standard convolution into depthwise convolutions and pointwise  convolutions. In depthwise convolution, single filter ($3\times 3$) is used for each input feature map  whereas pointwise convolution uses $1\times1$ filters to linearly combine the output feature maps of depthwise convolutions. Factorizing the standard convolution into two stages enables separating the  filtering and combining operations. This reduces both the number of parameters and MACs. 
 
\textbf{Limitations:}  
% From kernel analysis  (Table \ref{tab:Kernelanalysis}) we find that Gemv2T kernel comprises 55.5$\%$ of inference time (batch size one) and Gemv2N kernel comprise 30.5$\%$ of inference time. 
%As we show in Section \ref{sec:ExperimentalResults},
In MobileNet, $1\times 1$ convolution operation is responsible for 94.9$\%$ of total MACs  \cite{Mobilenet}. {\em Although the $1\times 1$ convolution does not require reordering in memory, thus no im2col overhead, it has low data reuse (compare to $3\times 3$ filter). Low data reuse implies low arithmetic intensity and high bandwidth requirement. This makes MobileNets energy inefficient} (Table \ref{tab:ResultsSummary}). 

As shown in Table \ref{tab:ResultsSummary},  for MobileNet, {\tt Gemv2T} and {\tt Gemv2N} kernels account for 55.5\% and 30.5\% of inference time, respectively.  Hence, MobileNet has stalls due to both memory dependencies and instruction dependencies. By using larger batch size, memory dependencies can be mitigated but instruction dependencies cannot be mitigated. This is evident from the fact that the inference time with $B=1$ and $B=90$ is 29.4ms and 23.6ms, respectively. Higher fraction of {\tt Gemv2T} and {\tt Gemv2N} kernels in total inference time leads to poor utilization of streaming multiprocessors (SMs). This further aggravates the low data re-usability problem of $1\times 1$ convolution.  These MACs with low data re-usability and poor SMs utilization  result into a throughput of 42 FPS  (frames-per-second), which is lowest among all the compact DNNs studied in this paper. 
 
% (Table \ref{tab:KernelPropertiesAnalysis}), which can be overcome using larger batch size and instruction dependencies (instruction fetch dependencies in Table \ref{tab:KernelPropertiesAnalysis}) which can't overcome by using larger batch size. 

\subsection{DenseNet}
DenseNet \cite{DenseNet} comprises of dense blocks and transition layers. 
In dense blocks, each layer is connected to  all the previous layers and produce same sized output feature maps.  In transition layers, the size of output feature maps reduces due to use of  convolution and pooling. Dense blocks have dense connectivity between layers such that each layer can access the output feature maps of all the previous layers. 
%i.e network's collective knowledge. 
%which  facilitates easy training, 
DenseNet reduces the number of parameters in following two ways. (1) As each layer in the dense block can access the  collective knowledge of the network, it does not require larger number of filters. Specifically, number of filters in dense blocks of DenseNet-121 are 32 and 128  whereas between 96 and 384 in AlexNet.  (2) DenseNet introduces the bottleneck layer to reduce the number of feature maps in transition layers. 

\textbf{Limitations:} Because of the dense connectivity in dense blocks, it requires multiple copies of the feature maps in the memory. Due to this, among the compact DNNs studied here, DenseNet has the highest memory   footprint of 1405MB, as shown in Table  \ref{tab:ResultsSummary}. {\em With a batch size of only 20,  it consumes nearly entire global memory of GPU P100.}    

\subsection{GoogLeNet and Inception-V2}
GoogLeNet \cite{GoogleNet} has $12\times $ fewer parameters than AlexNet but is significantly more accurate than AlexNet. It uses inception module as the building block and $1\times1$ filters as the bottleneck layer inside the inception module. This bottleneck layer reduces the dimension of input feature map and  hence, the number of MACs, before $3\times3$ and $5\times5$ convolutions. This technique allows  increasing the depth and width of network, which makes the  network more expressive and discriminative, without increasing the  computational budget. Similarly,  Inception-V2 \cite{InceptionV2} uses inception module in more aggressive manner and  factorizes the $5\times 5$ convolution into $3\times 3$ and further into $1\times 3$ and $3\times 1$ convolutions. This  reduces the number of parameters and  MACs. 

\textbf{Limitations:} Although factorizing the larger filter into smaller filters reduces computation and parameters but it increases the number of activations, as shown in Figure \ref{fig:ParameterActivationCalculation}. This increase in activations and decrease in MACs lowers MACs/activation ratio significantly and hence the arithmetic intensity which in turn increases bandwidth demand and reduces energy efficiency. This also makes Inception-V2  unsuitable for larger batch size. For example, on P100 GPU, GoogLeNet can work with a batch size of 128 whereas Inception-V2 cannot work due to exceeding the global memory size limitation.

\section{Experimental Results}
\subsection{Methodology}
%We perform our experiments \NJ{how much of the results are obtained from GPU and how much by mathematical analysis only} on . BVLC (Berkeley Vision and Learning Center)  \cite{P100Reference}

We run the DNNs using   Caffe framework \cite{caffe} on Tesla P100 GPU. P100 GPU has a global memory capacity of 12193 MB. Since P100 is a high-end GPU with ample amount of compute and memory resources, this by itself does not present any resource bottlenecks, and hence, we can focus on the characteristics of DNN itself, for example, we can find the performance bottlenecks of a DNN with large batch size.  
%The use of GPU is motivated by two reasons: (1) The massive parallelism provided by GPU makes it suitable for running  
% We have used GPU for our experiments because (1) It has very high degree of parallelism which is suitable for embarrassingly parallel applications (e.g. convolution in DNNs). (2) It has high compute (cuda cores) and memory (DRAM) resources which helps analyzing the actual performance bottlenecks of a DNN with larger batch size (because there will not be any resource constraints).    
We use CUDA version 8.0 and cuDNN version 7.0.1. For measuring the memory footprint and average power measurement of GPU,  we use \textit{nvidia-smi} utility. We record 50 samples per second. For measuring the performance of individual kernels, we use Nvidia visual profiler.

\textbf{Metrics:} We measure the correlation between different values using the Pearson product-moment correlation coefficient (PPMCC). PPMCC varies between +1 and -1 where values close to +1 or -1 indicate high degree of positive or negative (respectively) linear correlation and  a value of  0 indicates no linear correlation. 
For getting insights into  characteristics of GPU kernels, we use the following metrics \cite{P100Reference,ref77}. {\em Compute utilization } is combined utilization of arithmetic units, control-flow units, load-store units and functional units.  {\em Bandwidth utilization} shows the percentage utilization of the global memory bandwidth. \textit{Occupancy} shows the ratio of total number of active warps and the maximum warps supported in a streaming multiprocessor (SM). {\em Warp execution efficiency} is the average percentage of active threads in each executed warp. 

\subsection{Results} \label{sec:ExperimentalResults}
%{\em Notice that energy efficiency of most of the compact DNNs is quite low compared to AlexNet.}

\begin{table*} [htbp]
\caption{Inference time (ms), memory footprint (MB), energy efficiency ($ 10^9\times MACs/Joule$) and throughput (FPS). The last three columns show the percentage contribution of kernels {\tt Gemv2T}, {\tt Gemv2N} and {\tt Gemmk1} in the total inference time. }
\label{tab:ResultsSummary}
\centering
\begin{tabular}{ |c|c|C{2.2cm}|c|C{1.5cm}||c|c|c|} 
 \hline
 \textbf{Model Name} & \textbf{Inference time} & \textbf{Memory footprint}  &  \textbf{Energy efficiency} & \textbf{Throughput} & \textbf{Gemv2T($\%$)} &\textbf{Gemv2N($\%$)}& \textbf{Gemmk1($\%$)} \\
 \hline
     AlexNet        & 2.1     & 1015    & 75.94  & 4000 &  7.54  & 0.18  & 11.43 \\  
    \hline
    SqueezeNet-V1.0 & 3.8     & 615     & 40.95  & 1479 &   0    & 0     & 0             \\
    
    SqueezeNet-V1.1  & 3.5    & 587     & 27.07   & 2778 &   0    & 0    &  0  \\
   \hline
    1.0-G-SqNxt-23   & 24.4   & 1019    & 1.69   & 763  &  40.92 & 5.41  & 8.64           \\
    
    1.0-SqNxt-23     & 24.5   & 885     & 2.16   & 770 &  42.13 & 5.7   & 9.35           \\
    
    1.0-SqNxt-23v5   & 23.8   & 867     & 1.92   & 1004 &  31.68 & 6.82  & 10.73 \\
    
    2.0-SqNxt-23     & 28.2   & 995    & 3.75   & 412 &  35.10 & 5.06  & 7.99           \\
    
    2.0-SqNxt-23v5    & 27.7  & 957     & 3.86   & 541 &  24.71 & 6.53  & 8.61           \\
    \hline
     1.0-MobileNet-224& 29.4  & 733     & 1.65   & 42 &  \textbf{55.47} & \textbf{30.5}  & 0.68 \\
    
    DenseNet-121      & 33.0  & 1405   & 8.41  & 182 &  10.04 &  0   & 5.34           \\
    \hline
    GoogLeNet         & 11.2  & 801    & 21.95   & 1333 &  0.22  & 0.16  & 0.05           \\
    
    Inception-V2      & 19.0  & 987    & 14.05   & 658 &  6.38  & 0.03  & 3.98           \\
    \hline
\end{tabular}
\end{table*}

Table \ref{tab:ResultsSummary} summarizes  the results on  memory footprint, latency, throughput and energy efficiency. Also it illustrates the percentage contribution of kernels in total inference time. We analyze these results to get deeper insights. Table  \ref{tab:Modelattributes} shows the number of parameters and  activations of various DNNs. Notice that {\em except AlexNet, all the DNNs have more activations than parameters}. AlexNet has 61M parameters but only 2.05M activations whereas 1.0-G-SqNxt-23, which is a variant of SqueezeNext, has 0.54M parameters but 17.81M activations. Also, DensenNet-121 has 8M parameters and 70M  activations. Evidently, {\em compact DNNs have large disparity in number of parameters and activations}.

% \begin{figure}[htbp]
% \fbox{\includegraphics[scale=0.28]{Figure/act_param_comp.png}}
% \caption{Comparison of number of parameters and number of activations for batch size one. }
% \label{fig:ParameterActivationComparison}
% \end{figure}

As shown in Table \ref{tab:PPMCCfootprint}, there is a high linear correlation between number of activations and memory footprint. This substantiates our claims that (1) the memory footprint depends on both the number of parameters and activations (2) the contribution of activations in memory footprint is significantly higher than that of parameters. Hence, we recommend that {\em future techniques for designing compact DNNs should also focus on reducing the number of activations}.

\begin{table}[htbp]
\caption{Correlation between values}
\label{tab:PPMCCfootprint}
\centering
\begin{tabular}{ |c|c|c| } 
 \hline
 \textbf{$X$}& \textbf{$Y$} &\textbf{PPMCC($X$,$Y$)}\\ \hline
 $\#$ Parameters     & Memory footprint & 0.24 \\
 $\#$ Activations               &Memory footprint & 0.75 \\
 $\#$ Parameters + $\#$ activations                 & Memory footprint & 0.82 \\
 \hline

\end{tabular} 
\end{table}

\begin{figure}[htbp]
\includegraphics[scale=0.35]{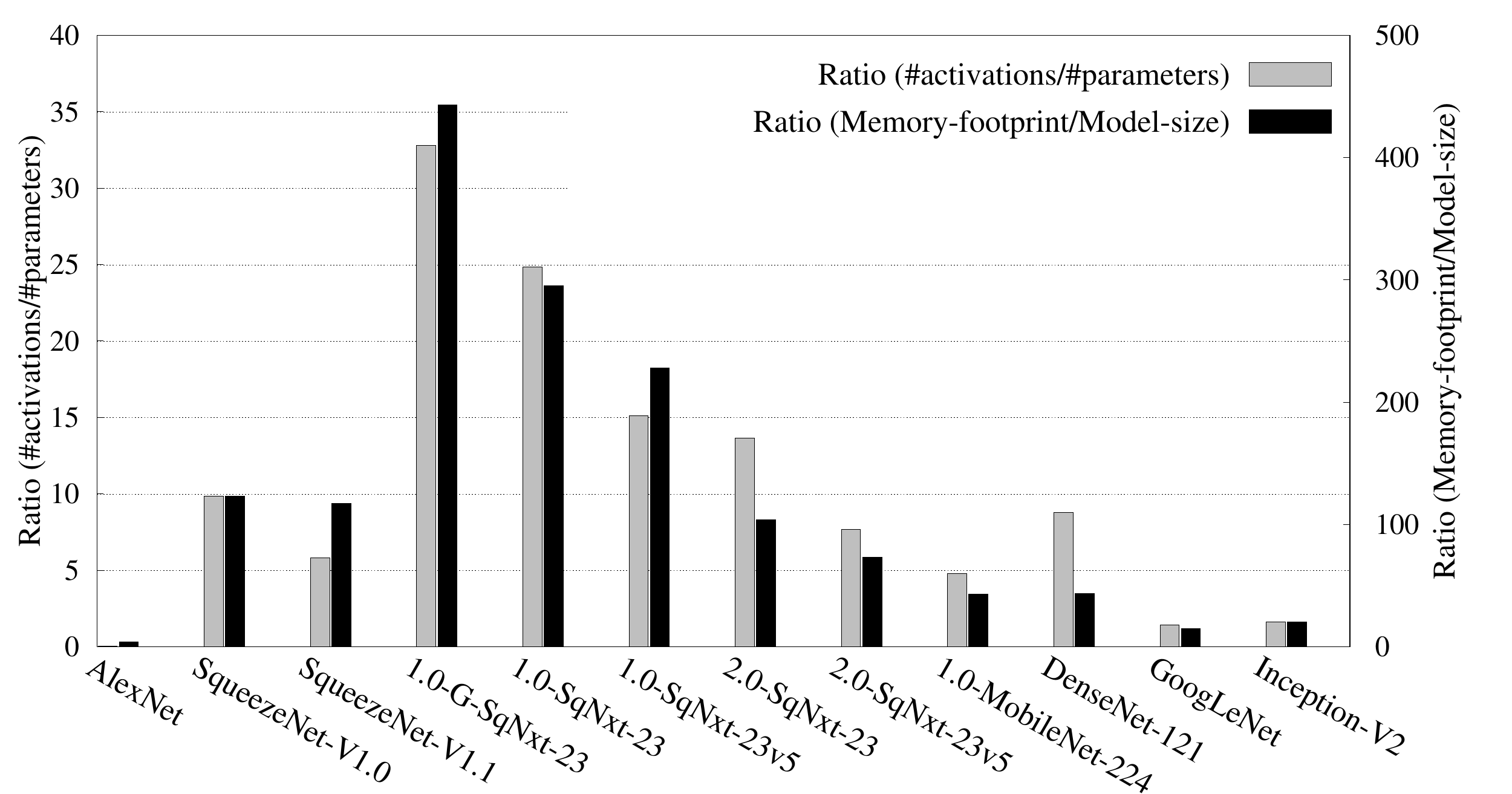}
\caption{Activations/Parameters and 
 memory-footprint/model-size ratio. Lower value of both ratios is better.}
\label{fig:RatiosComparison}
\end{figure}

Figure \ref{fig:RatiosComparison} shows activations to parameters ratio and memory footprint to model size ratio. Evidently,  as the ratio of activations to parameters (left vertical axis) increases, the ratio of total memory footprint to model size (right vertical axis) also increases. Compared to AlexNet, which has activations  to parameters ratio 0.03 and total memory-footprint to model size ratio 4.16,  1.0-G-SqNxt-23 has activations  to parameters ratio of 32.8 and total memory-footprint to model size ratio of 443. The PPMCC between these two ratios  is 0.96 and thus, these two ratios are  strongly correlated.

%Higher activation to parameter ratio implies higher memory footprint (Global memory consumption in GPU with batch size=1) to model size ratio. Clearly, activation memory plays a vital role in the total memory footprint. 

\begin{figure}[htbp]
\includegraphics[scale=0.33]{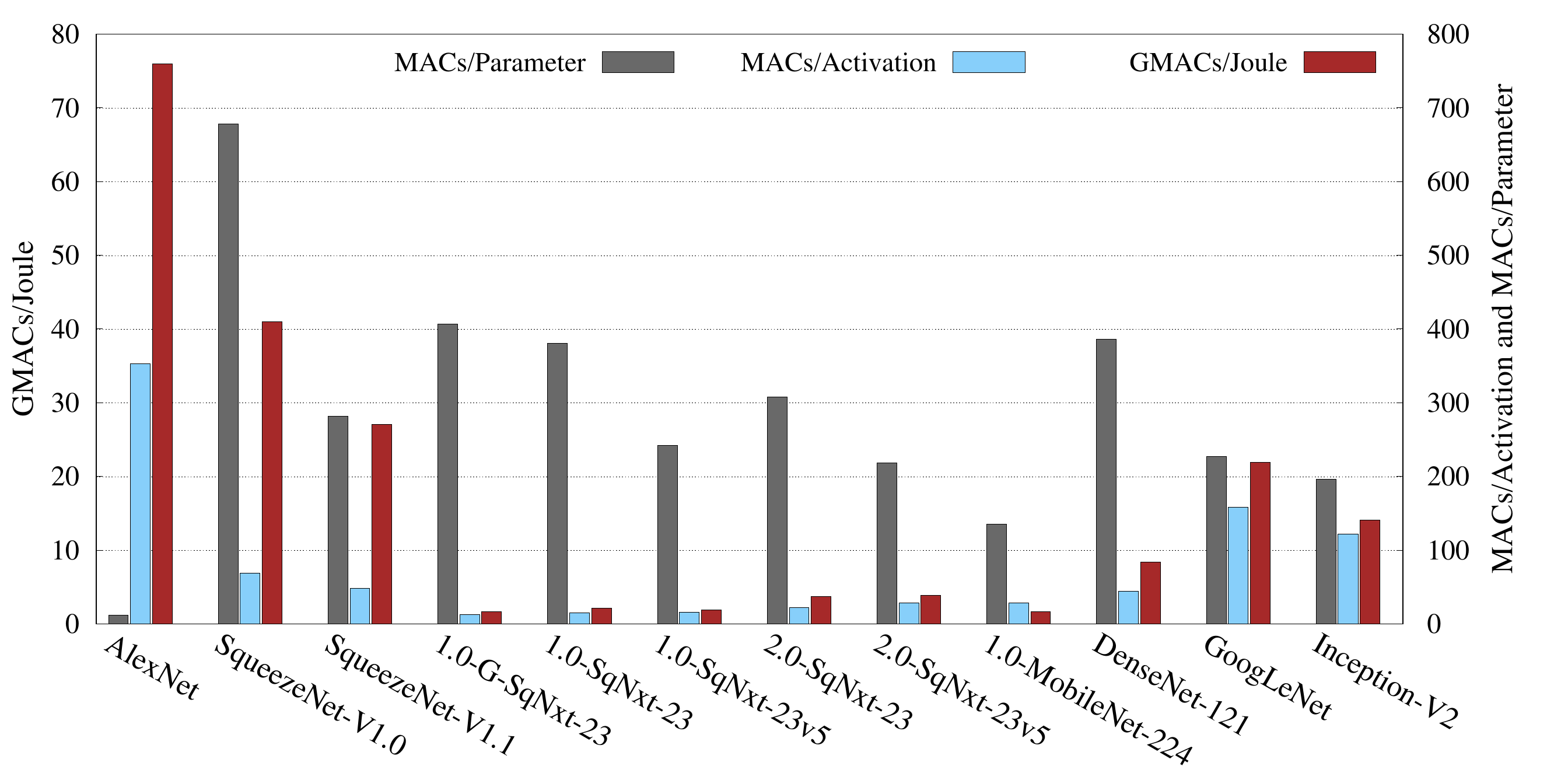}
\caption{Energy efficiency (left vertical axis) and MACs/parameter and MACs/activation (right vertical axis). For all three metrics, higher is better. \textit{( better viewed in color)}.
} 
\label{fig:EnergyComparison}
\end{figure}

We have used batch size four, to utilize the GPU compute resources, in calculation of energy efficiency.
Figure \ref{fig:EnergyComparison} shows the energy efficiency (left vertical axis), in terms of $MACs\times 10^9/Joule$. Clearly, energy efficiency depends not only on MACs/parameters ratio (right vertical axis) but also on the MACs/activations ratio (right vertical axis). 
%To find the linear correlation between the proxies for arithmetic/operational intensity and energy efficiency we calculate PPMCC. 
\textit{The PPMCC between energy efficiency and MACs/parameter ratio is -0.18 and that between energy efficiency and MACs/activation ratio is 0.88}. This shows that there is high linear correlation between MACs/activation ratio and energy efficiency. Therefore networks with comparable MACs/parameter ratio, but  higher  MACs/activation ratio have higher energy efficiency, for example 1.0-G-SqNxt-23 and 1.0-SqNxt-23. MACs/activation ratio of AlexNet is higher than that of all the compact DNNs studied here and  this makes AlexNet the  most energy efficient. Here, we recommend that {\em future techniques for designing energy efficient DNNs should focus on increasing the arithmetic  intensity, primarily by increasing the the MACs/activation ratio.}   

%When MACs/activation is higher then energy efficiency is also higher, provided their MACs/parameter is comparable.

\textbf{Kernel-level analysis:} 
To gain more insights, we perform kernel level analysis by profiling the DNNs on Tesla P100 GPU. We focus on three cuBLAS kernels, viz. {\tt Gemv2T}, {\tt Gemv2N}, {\tt Gemmk1} because these kernels present in  all the DNNs, except SqueezeNet, we analyzed in this paper. Table \ref{tab:KernelPropertiesAnalysis} shows the resource utilization of these kernels along with stall reasons. 
 Note that Gemm (general matrix-matrix multiplication) is a dense matrix style of computation and because of better data reuse \cite{Notallops},  it has  high arithmetic intensity. Gemv (general matrix-vector multiplication) is a sparse matrix style of computation and because of poor data reuse, it has lower arithmetic intensity.  The letters `T' in {\tt Gemv2T} and `N' in {\tt Gemv2N} stand for transpose and not-transpose respectively. Gemm-style  computation is faster than Gemv style \cite{Notallops} and we found that the {\tt Gemmk1} kernel, which is based on Gemm style computation,  has excellent SM utilization (Table \ref{tab:KernelPropertiesAnalysis}).

 %and which factors lead to throughput bottlenecks
%Throughput of a DNN depends  primarily  on the number of operations required to process inputs, i.e. number of MACs. 
%As shown in section \ref{sec:introduction}, MACs in larger filter convolution have higher data reuse which leads to higher arithmetic/operational intensity. 

%Our kernel-level analysis provides insights into how well the  MACs are utilizing the compute resources of GPU. 

%which is helpful in understanding the throughput bottlenecks. 

\begin{table}[htbp]
\caption{Analysis of kernel properties}
\label{tab:KernelPropertiesAnalysis}
\centering
\begin{tabular}{ |p{3.4cm}|p{1.1cm}|p{1.2cm}|p{1.1cm}| } 
 \hline
 \textbf{Attributes/stall reasons}& \textbf{Gemv2T} &\textbf{Gemv2N}& \textbf{Gemmk1}\\
 \hline
 Compute utilization  ($\%$)    & 15     & 8        & 5        \\
 Bandwidth utilization ($\%$)   & 0.92   & 0.094    & 11       \\
 Occupancy ($\%$)               & 6.2    & 6.2      & 11.8     \\
 Warp execution efficiency($\%$)& 80.7   & 99.3     & 98.4     \\
 Branch divergence ($\%$)       & 25     & 0        & 4        \\
 \textbf{SM utilization}        & \textbf{Poor}   & \textbf{Very poor}& \textbf{Excellent} \\
 \hline
 \textbf{Memory dependency ($\%$)}& \textbf{44.3}   & \textbf{3.6}  &\textbf{54.4} \\
 %Execution dependency ($\%$)     & 16     & 10.7     & 5.4   \\
 %Instruction issue ($\%$)       & 5.2    & -        & 1.8   \\
 %Instruction fetch ($\%$)       & 10.8   & 46.2     &11.8   \\
 \textbf{Instruction dependency ($\%$)}  & \textbf{32} & \textbf{56.9}  &\textbf{19} \\
 Synchronization ($\%$)         & 8.8    & 25       & 9.1   \\
 Others ($\%$)                  & 15     & 14.5     & 17.5  \\
 \hline
\end{tabular} 
\end{table}

%To ascertain the exact bottleneck, we look closely at the stall reasons. 

Both the compute utilization and bandwidth utilization of these three kernels is well below 60\% and hence,  the performance of these kernels is bounded by either instruction latency  or memory latency. In  {\tt Gemv2T} and  {\tt Gemmk1}, memory dependency causes 44\% and 54\% of total stalls and hence, these kernels are bottlenecked by memory latency. Also, high branch divergence (25\%) in {\tt Gemv2T} reduces its warp execution efficiency which further reduces the utilization of SMs. For  {\tt Gemv2N}, instruction dependency causes 57\% of total stalls and thus, it is bottlenecked by instruction latency. 
%Note that instruction dependency is the sum of execution dependency, instruction issue stall and instruction fetch stall.  

It is well-known that the memory latency bottleneck can be mitigated by using larger batch size, whereas the  instruction latency bottleneck can be mitigated only by higher instruction-level parallelism. A high value of instruction dependency leads to poor utilization of SMs. Hence, the SMs utilization of {\tt Gemv2N} kernel is very poor. We conclude that a network, which spends high fraction of time in {\tt Gemmk1} and low fraction of time in {\tt Gemv2T} and {\tt Gemv2N}, will show higher utilization of SMs. For example, MobileNet has $5.4\times$ fewer MACs than DenseNet (Table \ref{tab:Modelattributes}) but $4.3\times$ lower throughput (Table \ref{tab:ResultsSummary}). This is because {\tt Gemv2T} and {\tt Gemv2N} together contribute 86\% of inference time in MobileNet but only 10\% of  time in DenseNet (Table \ref{tab:ResultsSummary}). 
 
%Table \ref{tab:ResultsSummary} shows the percentage contribution  of kernels Gemv2T, Gemv2N and Gemmk1 in the total inference time. 

%\NJ{Add some comment, such as compared to AlexNet, other networks have ..... }

%\begin{table}[htbp]
%\caption{Percentage contribution of Gemv2T, Gemv2N and Gemmk1 kernels in total inference time. }
%\label{tab:Kernelanalysis}
%\centering
%\begin{tabular}{ p{2.1cm}|p{1.3cm}|p{1.3cm}|p{1.3cm} } 
% \hline
% \textbf{Model Name}& \textbf{Gemv2T($\%$)} &\textbf{Gemv2N($\%$)}& \textbf{Gemmk1($\%$)}\\
% \hline
% AlexNet          &  7.54  & 0.18  & 11.43          \\
% SqueezeNet V1.0  &   -    & -     & -              \\
% SqueezeNet V1.1  &   -    & -     &  -             \\
% 1.0-G-SqNxt-23   &  40.92 & 5.41  & 8.64           \\
% 1.0-SqNxt-23     &  42.13 & 5.7   & 9.35           \\
% 1.0-SqNxt-23v5   &  31.68 & 6.82  & 10.73          \\
% 2.0-SqNxt-23     &  35.10 & 5.06  & 7.99           \\
% 2.0-SqNxt-23v5   &  24.71 & 6.53  & 8.61           \\
% 1.0MobileNet-224 &  \textbf{55.47} & \textbf{30.5}  & 0.68           \\
% DenseNet-121     &  10.04 &  -    & 5.34           \\
% GoogleNet        &  0.22  & 0.16  & 0.05           \\
% Inception-V2     &  6.38  & 0.03  & 3.98           \\
% \hline
%\end{tabular}
%\end{table}

%\NJ{I suggest you remove the following lines from the caption: Lower percentage of Gemv2T and Gemv2N, higher percentage of Gemmk1 is better.}

%(from inter layer data dependency standpoint)
 
\subsection{Discussion} \label{sec:discussion}
In linear DNNs, such as AlexNet and VGGNet,  neighboring layers are connected in sequential order and producer-consumer relationships exist only between a layer and its immediate neighboring layers \cite{Superneuron}. By comparison, in non-linear DNNs, such as SqueezeNet, SqueezeNext, DenseNet etc., producer-consumer relationships exist between a layer and more than one of its neighboring layers. 
%This nonlinearity makes the memory over-provisioning problem of network wide memory allocation policy \cite{Superneuron,vDNN} worse.
This nonlinearity aggravates the problem of memory over-provisioning  \cite{Superneuron}. We found that activations/parameters ratio for linear DNNs is low, e.g., for AlexNet, it is only 0.03. By comparison, for all compact DNNs studied in this paper, the ratio is higher than one. {\em We propose using ``activation to parameter ratio'' as the measure of nonlinearity of the DNN}. 

In what follows, we classify the compact DNNs in groups and discuss the implications of architectural characteristics on memory footprint, energy efficiency and throughput.
  %, i.e., after conv1, fire3 and fire5, whereas in the baseline it is after conv1, fire4 and fire8 layers. This
  
\textbf{SqueezeNet family:} Compared to the baseline V1.0, V1.1 has two changes \cite{SqueezeNet}. First, it has low dimensional filters in conv1 layer (64, $3\times 3$ filters) compared to the baseline (96,$7\times 7$ filters). This reduces the number of MACs and parameters by 87\% while the number of activations are reduced by only 31\% in the conv1 layer. Second, it introduces pooling layers earlier \cite{SqueezeNet}  which reduces the size of output feature map and hence, the number of activations. As shown in Table \ref{tab:Modelattributes}, these  changes (1) decrease the number of MACs  leading to higher throughput, (2) decrease the number of activations leading to lower memory footprint, and (3) have negligible impact on the number of parameters. %However, the energy metrics (MACs/activations and MACs/parameters) are higher than that of the baseline and  it has lower energy efficiency (MACs/Joule).
However, the value of metrics viz., MACs/activation and MACs/parameter are decreased and hence,  V1.1 has lower energy efficiency than V1.0.
%Clearly, {\em compared to V1.0, V1.1 has higher throughput but lower energy efficiency.} 

%Clearly, {\em reducing MACs by $2.2\times$ in SqueezeNetV1.1 compared to baseline increases the throughput but reduces energy efficiency.}
 
 %1.0-SqNxt-23v5 and 2.0-SqNxt-23v5 
 
\textbf{SqueezeNext family:} The v5 versions of SqueezeNext have lower number of blocks in groups with higher input resolution and higher number of block in groups with lower input resolution. Although this change increases the number of parameters but reduces the number of activations hence  memory footprint. Also, this reduces the number of MACs and percentage contribution of { \tt Gemv2T} kernel in total inference time, which improves SMs utilization. Both these factors increase the throughput (Table \ref{tab:ResultsSummary}). Increasing the number of blocks in a group with lower input resolution and decreasing the number of blocks in groups with higher input resolution reduces MACs/parameter but increases MACs/activation. In 1.0-SqNxt-23v5, the increase in MACs/activation is very small which does not make it more energy efficient than 1.0-SqNxt-23. By contrast, 2.0-SqNxt-23v5 has higher MACs/activation than 2.0-SqNxt-23 which   increases its energy efficiency.

%As we describe in section \ref{sec:SqNxt} higher number of blocks in a group with higher input resolution leads to higher number of activations and it lowers the arithmetic/operational intensity. So to make arithmetic/operational intensity higher (1) there should be lower number of blocks in a group with higher input resolution i.e. closer to input and (2) there should be higher number of block in group with lower input resolution i.e. farther from input. This leads to higher MACs/activations, lower memory footprint and higher throughput. As there is a small increase  in GMAC/activations in v5 version of 1.0-SqNxt-23 and reduction  in MACs/parameters, the energy efficiency is reduced. However, in case of v5 version of 2.0-SqNxt-23, this increase is significant which  leads to higher energy efficiency. 

\textbf{Inception module, DenseNet and MobileNet:} We compare inception module family (GoogLeNet and Inception-V2) with DenseNet and MobileNet. Among these, MobileNet and Inception-V2 have lowest and highest number of parameters, respectively. Also, MobileNet and DenseNet have the lowest and highest memory footprint, respectively. This is because DenseNet has the highest number of  activations. DenseNet has highest number of activations due to its densely  connected structure which leads to multiple copies of the same  activation maps during the processing of inputs and it exponentially increases  with larger batch size. 
%{\em With a batch size  of only 20, it consumes almost all the global memory in P100 GPU.}

Further, among these DNNs, MobileNet has lowest number of MACs, so one may expect it to have the highest energy efficiency. But, {\em MobileNet is in fact the least energy efficient} due to (1) poor data reuse and (2) very high fraction of {\tt Gemv2T} and {\tt Gemv2N} kernels in total inference time (Table \ref{tab:ResultsSummary}). So MACs in MobileNet utilizing  SMs very poorly. For the same reason, MobileNet also has the lowest throughput. 

%high fraction of Gemv2T and Gemv2N kernel results into poor utilization of SMs and also leads lowest throughput of MobileNet. 

   %(e.g. SqueezeNet family, SqueezeNext family, Inception module family and rest)

%(to reduce the number of parameters/MACs) 
%
%Machine learning frameworks such as caffe and MXNet uses network wide memory allocation techniques where all the data required to process the inputs , such as activations and filter parameters , put into global memory of GPU. The main reason for this network wide memory allocation in GPU is to exploit performance benefits. Page migration overhead (from CPU to GPU) leads to under utilization of PCIe bandwidth (16 GB/s) and this becomes more pronounced when amount of data to be transferred from CPU to GPU is in order of GBs \cite{vDNN}. Whereas {\tt cudaMemcpy} achieves an average data transfer speed of 12.8 GB/s.

\section{Conclusion and Future work} \label{conclusion}
In this paper, we have analyzed the state-of-the art compact DNNs and the implications of parameter-reducing techniques on memory footprint, energy efficiency and throughput. We show that there is very high correlation between number of activations and memory footprint. Naively reducing number of parameters can lead to very high number of activations which increases the memory footprint. Also, the activations/parameters ratio is representative of memory-footprint/model-size ratio. Lower number of MACs does not necessarily improve energy efficiency and throughput. Energy efficiency also depends on arithmetic  intensity which we estimate in terms of MACs/parameter and MACs/activation. Throughput of these compact DNNs depends on the number of MACs as well as how these MACs are utilizing the SMs. 
%This utilization depends on the percentage contribution of {\tt Gemmv2T}, Gemv2N and Gemmk1 kernels. 
We also provide important guidelines for designing compact and energy efficient DNNs: while reducing number of parameters and MACs, researchers should also ensure high arithmetic and operational intensity  and  low activations to parameters ratio. 
Our future work will focus on optimizing DNNs on different hardware platforms \cite{ref98} and developing tools to estimate the energy efficiency of the DNNs at design time.

%, on general purpose hardware (e.g. CPU, GPU etc.). This would help the researchers to design the DNNs according to the compute, memory and energy capability of given platforms.
%which leads to high energy efficiency, , which leads to low memory footprint

%\NJ{Cite not CoRR but the conferences}
 
%\input{20acknowledgment}
{
\scriptsize
 \linespread{0.97}
\bibliographystyle{IEEEtran1}
\bibliography{Ref}
}

%\section*{References}

\end{document}